\pdfoutput=1

\documentclass[11pt]{article}

\usepackage[]{acl}

\usepackage{times}
\usepackage{latexsym}
\usepackage{booktabs}
\usepackage{hyperref}
\usepackage{graphicx}
\usepackage{mdframed}

\usepackage[T1]{fontenc}

\usepackage[utf8]{inputenc}

\usepackage{microtype}

\usepackage{inconsolata}

\title{Unsupervised Domain Adaption for Neural Information Retrieval}

\author{Carlos Dominguez \and Jon Ander Campos \and Eneko Agirre \and Gorka Azkune \\
  HiTZ Basque Center for Language Technologies - Ixa NLP Group \\
  University of the Basque Country (UPV/EHU) \\
  M. Lardizabal 1, Donostia 20018, Basque Country, Spain \\
  \texttt{\{carlos.dominguez, jonander.campos, e.agirre, gorka.azcune\}@ehu.eus} \\
}
\begin{document}
\maketitle
\begin{abstract}
Neural information retrieval requires costly annotated data for each target domain to be competitive. Synthetic annotation by query generation using Large Language Models or rule-based string manipulation has been proposed as an alternative, but their relative merits have not been analysed. 
In this paper, we compare both methods head-to-head using the same neural IR architecture. We focus on the BEIR benchmark, which includes test datasets from several domains with no training data, and explore two scenarios: zero-shot, where the supervised system is trained in a large out-of-domain dataset (MS-MARCO); and unsupervised domain adaptation, where, in addition to MS-MARCO, the system is fine-tuned in synthetic data from the target domain.  
Our results indicate that Large Language Models outperform rule-based methods in all scenarios by a large margin, and, more importantly, that unsupervised domain adaptation is effective compared to applying a supervised IR system in a zero-shot fashion. In addition we explore several sizes of open Large Language Models to generate synthetic data and find that a medium-sized model suffices. Code and models are publicly available for reproducibility.
\end{abstract}

\section{Introduction}
In recent years, there has been significant advancement in Information Retrieval (IR). IR is the process of retrieving relevant information from a collection of unstructured or semi-structured text data in response to a user's information need. With the explosion of digital content in recent years, IR has become an increasingly important field of research with applications in search engines \cite{ir_for_search_engines}, recommendation systems \cite{recommender_systems}, chatbots \cite{chatbots}, and more.

\begin{figure}[t] 
  \begin{center}
    \includegraphics[width=8cm]{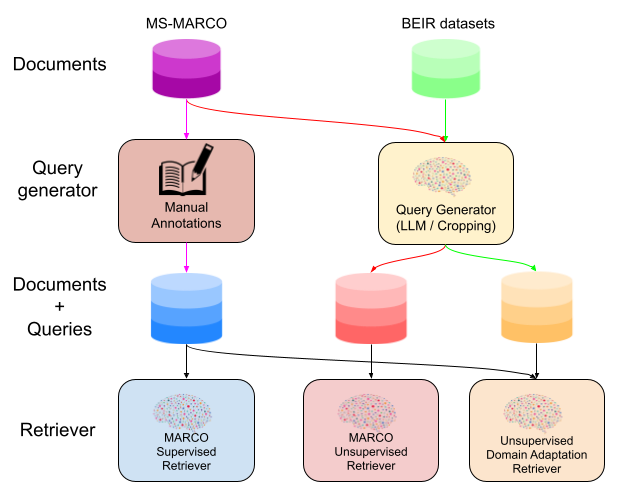} 
    \caption{Experimental design: (left) a supervised retriever is trained with manual annotations from MS-MARCO; (middle) an unsupervised retriever is trained with automatically generated queries for MS-MARCO documents; (right) an unsupervised domain adaptation retriever is trained with both MS-MARCO manual annotations and automatically generated queries in-domain BEIR dataset documents. Evaluation is performed in BEIR producing two scenarios: zero-shot (left and middle retrievers); unsupervised domain adaptation (right retriever).}
    \label{fig:pipeline} 
  \end{center} 
\end{figure}

Traditionally, IR systems used statistical models, such as BM25 \cite{BM25}, to represent documents and queries and measure their similarity. However, these models have limitations, such as the inability to capture semantic relationships between words, the requirement of exact matches between query and documents tokens - vocabulary mismatch \cite{vocabulary_mismatch}), and the lack of interpretability.

In recent years, deep learning techniques, such as dual encoders, have emerged as a promising approach for IR. Dual encoders use two neural networks, one to encode the query and the other to encode the document, into high-dimensional vector representations. The similarity between the two vectors is then computed to rank the documents based on their relevance to the query. 

Dual encoders have several advantages over traditional IR models. They can capture complex semantic relationships between words and phrases and learn from large-scale data, making them more adaptable to different domains. Additionally, they can be fine-tuned for specific tasks and easily integrated into existing systems. 

Recent studies have shown the effectiveness of dual encoders in various IR tasks, including passage retrieval \cite{DPR}, question answering \cite{FiD}, and conversational search \cite{conversational_search}. For example, the Dense Passage Retrieval (DPR) \cite{DPR} , one of the most common systems, uses dual encoders based on the BERT architecture \cite{bert}. On the downside, neural IR models need to be trained on relatively large and costly datasets which include query and document pairs, e.g. MS-MARCO \cite{MSMARCO}.

In many real-world applications, the performance of machine learning models degrade when deployed in a new environment or when applied to data that differ from the training data. This phenomenon, known as domain shift, can lead to poor generalization and decreased effectiveness. Domain adaptation (DA) techniques adapt a model that has been trained on one domain (or dataset) to work effectively on a different but related domain. DA is effective whenever there is a mismatch between the distribution of data on which a model was trained and the distribution of data on which it will be used in practice.

In information retrieval, out-of-domain datasets usually lack the necessary annotations to support supervised domain adaptation. This makes it challenging to train models that can be effectively deployed in new domains. To overcome these challenges, unsupervised techniques to generate synthetic training data have become increasingly popular in recent years. The main idea is to take unlabeled documents from the domain and generate a query (or queries) for each document, resulting in an automatically annotated dataset. The most used techniques range from rule-based systems such as the Contriever cropping system  \cite{CONTRIEVER} and the Inverse Cloze Task from LaPraDoR \cite{laprador}, or leveraging generative Large Language Models (LLMs) to generate queries, such as Promptagator \cite{promptagator} and InPars \cite{INPARS}. Table \ref{contriever-examples} shows two examples of the synthetic data generated by the a LLM (left part) and a cropping rule-based method (right part). 

Although different question generation have been proposed, they have not been evaluated head-to-head, that is, we do not know which one is the best. In this paper, we propose a controlled evaluation setting to compare unsupervised query generation systems on equal terms, specially focused on a domain adaptation scenario. Our evaluation set-up is based on a well-known neural IR system (SBERT \cite{sbert}), which is kept as a fixed variable in all the experiments, and focuses solely on the impact of query generation systems. We follow standard practice and use MS-MARCO to train the base IR system \cite{MSMARCO}, as well as the BEIR benchmark for evaluation\cite{BEIR}. The methodological design (see Figure \ref{fig:pipeline}) has the following steps: (i) use a large annotated dataset (MS-MARCO) to train a base IR system; (ii) select a collection of out-of-domain IR datasets (BEIR); (iii) generate queries for the out-of-domain documents using different unsupervised approaches; (iv) fine-tune the base IR system with the synthetic queries and their corresponding documents for domain adaptation; (v) compare the performance using well-established metrics on BEIR. 

Our experiments show the following:

\begin{itemize} 
    \item LLM outperforms rule-based query generation in terms of retrieval performance, indicating the superior capabilities of LLM in generating effective queries.
    \item Unsupervised domain adaptation performance is improved by using LLM for generating queries, while the rule-based independent cropping system from Contriever fails.
    \item In addition, we analyse the effect of increasing the amount of parameters of the query generation LLM for the IR module performance.
\end{itemize}

\let\thefootnote\relax\footnotetext{For reproducibility, we perform experiments using open-source LLMs. The code is available at \href{https://github.com/CarlosDominguezBecerril/InformationRetrieval}{https://github.com/CarlosDominguezBecerril/InformationRetrieval}. }

\begin{table*}[t]
    \scriptsize
        \begin{tabular}{{p{4cm}p{1.8cm}|p{1.9cm}p{2.0cm}p{1.9cm}p{1.9cm}}}
        \toprule
         \bf Original Document &   \bf LLM   &  \multicolumn{4}{c}{\bf Rule-based string manipulation}  \\
        &   \bf Query   &   \textbf{Document} \newline (random cropping) &   \textbf{Final Document} (random cropping + words deletion)  &   \textbf{Query} \newline (random cropping)&   \textbf{Final Query} (random cropping + words deletion) \\        \midrule
        Color hex is a easy to use tool to get the color codes information including color models (RGB,HSL,HSV and CMYK), css and html color codes. & What is the color hex? &  
        get the color codes information including color models (RGB,HSL,HSV and CMYK) &  
        color codes information color models (RGB,HSL,HSV and CMYK) &
        Color hex is a easy to use tool &  
        Color hex tool   \\
        \midrule
        Although the European powers did make military interventions in Latin America from time to time after the Monroe Doctrine was announced, the Americans did not look for war. They did, however, use the doctrine as justification for taking Texas in 1842 under President John Tyler. &
        What was the Monroe Doctrine? &
        the Americans did not look for war. They did, however, use the doctrine as justification for taking Texas &  
        Americans look for war. the doctrine justification taking Texas &
        military interventions in Latin America from time to time after the Monroe Doctrine was announced &  
        military interventions in Latin after Monroe \\
        \bottomrule
        \end{tabular}
    
    \caption{Synthetic data (document and query) produced using two methods for two sample documents. The query generated by the LLM (OPT-2.7B) from each document is paired with the original document, i.e the synthetic dataset comprises the  two leftmost columns. The rule-based method generates both synthetic document and query for each original document in two steps: randomly crop the input and then delete random words. In this case The synthetic dataset comprises the fourth and sixth columns.}
    \label{contriever-examples} 
\end{table*}

\section{Related Work}

In this section, we will briefly review current IR systems and query generation methods.

\subsection{Retrieval models}
In the field of IR, documents and queries are typically represented as sparse vectors, with each element corresponding to a term in the vocabulary. BM25 \cite{BM25} is a well-known ranking function that ranks documents based on query terms within a document, without considering the relationship between query terms. BM25 is a family of scoring functions, with different components and parameters. On the other hand, Dense Passage Retrieval (DPR) \cite{DPR} and Sentence-BERT (SBERT) \cite{sbert} are retrieval methods that use a two-tower model architecture. The first encoder builds an index of all text passages, while the second encoder maps the input question to a vector and retrieves the top $k$ passages with the closest vectors. The similarity of the vectors is calculated by using the dot product or cosine similarity. Moreover, they optimize the negative log-likelihood loss function to create a vector space where relevant pairs of questions and passages have higher similarity than irrelevant ones, using in-batch negatives as negative passages. The two most important differences are: (i) SBERT uses tied encoders (shared weights), whereas DPR uses two independent encoders; (ii) SBERT uses mean pooling to obtain the final vector, while DPR makes use of the $[CLS]$ token. Modern IR models allow fine-grained token-level interaction to improve the performance but with higher inference cost. Two of such models are ColBERT \cite{colbert} and SPLADE \cite{splade}. The main difference between DPR and ColBERT is in their approach to encode the document and query representations. ColBERT uses a joint space approach and a late interaction strategy, while DPR uses a dual-encoder architecture and a dense retrieval approach.

\subsection{Query generation systems}
Query generation is a crucial aspect of IR that aims to generate high-quality queries automatically for synthetic training data generation. Among the various approaches, Large Language Models as used by Promptagator \cite{promptagator} and the rule-based independent cropping system from Contriever \cite{CONTRIEVER} are very effective widely used.

Promptagator \cite{promptagator} utilizes LLMs to generates task-specific queries by combining a prompt with a few examples from the domain (few-shot domain adaptation). The approach consists of two components: prompt-based query generation and consistency filtering. Prompt-based query generation is performed using FLAN, a LLM with 137 billion parameters not publicly available. The generated queries are then filtered based on consistency, that is, a query should be answered by the passage from which the query was generated.

Another system that makes use of LLMs is InPars \cite{INPARS}. This paper makes use of BM25 to retrieve the top 1000 candidate documents, and then a monoT5 \cite{monoT5} neural reranker is used to reorder the documents. Reranking is performed using unsupervised generated queries by GPT-3 \cite{GPT3}.

In contrast, Contriever \cite{CONTRIEVER} proposes a rule-based method that uses independent cropping as a data augmentation method for text. In addition it applies also word deletion, replacement, or masking. From now on, we are going to refer to the independent cropping from Contriever as \textit{independent cropping} or \textit{cropping} for short.

There exist several more techniques, such as using the Inverse Cloze Task (ICT) \cite{ICT}, used by LaPraDoR \cite{laprador}, where a passage is broken down into multiple sentences and one sentence is used as a query while the remaining ones are joined and used as the document. On the other hand, GenQ \cite{BEIR}, trains a T5 \cite{T5} model on MS-MARCO for 2 epochs and then generates 5 queries per document using different sampling methods. These two methods are not utilized in our study. GenQ is a supervised approach, and therefore, does not fall under the scope of our investigation. On the other hand, ICT has exhibited lower performance compared to independent cropping, making it unsuitable for our purposes.

The two highly promising techniques, Large Language Models (LLM) and rule-based methods, have been extensively employed either independently or in combination. Nevertheless, the question of which approach is more effective remains unresolved as no direct head-to-head comparisons have been conducted to date.

\section{Dataset Description and Generation}
\label{LLM_for_query_generation}

In this section, we discuss how we applied independent cropping and LLMs to generate questions for a given collection of unlabelled documents.

Contriever \cite{CONTRIEVER} proposes to use independent cropping when unlabelled data is supplied as training data. Independent cropping is a common independent data augmentation used for images where views are generated independently by cropping the input. In natural language processing, it would be equivalent to sampling a span of tokens. Furthermore, in addition to independent cropping, they also consider different data augmentation methods such as word deletion, replacement, or masking, being word deletion the one that works the best. Following their experiments, we apply independent cropping and word deletion to obtain query-document pairs.

On the other hand, LLMs can be used to generate questions in an unsupervised way (zero-shot question generation). In order to generate questions, we need to make use of a prompt. We selected the vanilla version of the prompts used by Inpars \cite{INPARS}. To include diversity in the generated questions, we apply sampling with a top $p$, that is only the smallest set of most probable tokens with probabilities that add up to top $p$ or higher are kept for generation. We set $p$ on 0.9 based on prior work and did not try any other value.

The Open Pretrained Transformer (OPT) \cite{OPT} is proposed as the model for this paper due to several reasons. Firstly, it is an open-source model. Secondly, it has a variety of checkpoints available, ranging from small (125 million parameters) to large models (175 billion parameters), enabling measuring scalability of parameters with the quality of generated queries. Lastly, our experimental results showed that OPT performed as well as other models such as Bloom \cite{BLOOM}, GPT-neoX \cite{gpt_neox}, and GPT-neo \cite{gpt_neo} making it a viable option for our study. Further details about the use of Large Language Models for query generation can be found in Appendix \ref{LLM_extra_info}.

Examples of queries generated using LLMs and independent cropping generation methods can be found in Table \ref{contriever-examples}. The table shows how query-document pairs are generated by each method. LLMs produce queries that are more akin to human language and are therefore simpler to comprehend. In contrast, independent cropping generates strings based on the content of the document. Based on our analysis, we believe that LLMs produce higher quality document-query pairs overall.

\section{Experimental setting}
In this section, we explain the models used for the experiments, the metrics used to measure the performance, and the benchmark used to compare the performance of the systems.

\subsection{Models: IR model}

Recent papers make use of bi-encoder architectures to train their systems. Two such techniques are SBERT and DPR which have shown promising results. We experimented with both, and found that SBERT outperforms DPR. Experiments comparing SBERT and DPR can be found in Appendix \ref{dpr_vs_sbert}.

Following the trend we train a two tower architecture system based on SBERT that makes use of a neural network to obtain embedding representations. This is done in two step: first the neural network matches query features $x_{query}$ to query embeddings $\psi(x_{query})$ and then it matches context features $x_{context}$ to context embeddings $\psi(x_{context})$. The output of the model can be defined as the cosine similarity of $<\psi(x_{query}), \psi(x_{context})>$, which can be seen as the similarity between the query and the context. Following the SBERT model, we also use DistilBERT \cite{distilbert} as the base model. Furthermore, we employ mean pooling to obtain the final vector representation, which has been shown  to outperform the use of the $[CLS]$ token on different Natural Language Processing downstream tasks \cite{cls_bad}.

Bi-encoder models are trained contrastively using in-batch negatives, that is, for each question in the batch, we will consider the positive passages of the other questions as the negative passages. For a batch of size $B$, we will have for each question 1 positive passage and $B-1$ negative passages.

To ensure reproducibility, the following hyperparameters were used to train the model: \textit{distilbert-base-uncased} was chosen as the base model instead of BERT due to its similar results in preliminary experiments and faster training and evaluation. The batch size was set to 256 and the number of epochs to 10. For model selection, the epoch with the highest nDCG@10 metric in the development dataset was chosen. In unsupervised DA, additional 1000 examples were generated to create the development dataset.

\subsection{BEIR Benchmark}
In this paper, we make use of \textbf{Be}nchmarking-\textbf{IR} (BEIR). BEIR is a robust and heterogeneous evaluation benchmark for IR, comprising 18 retrieval datasets for comparison and evaluation of model generalization. BEIR is focused on diversity, that is, the benchmark includes nine different retrieval tasks: fact checking, citation prediction, duplicate question retrieval, argument retrieval, news retrieval, question answering, tweet retrieval, bio-medical IR, and entity retrieval. Furthermore, datasets from diverse domains are included to cover broad topics like Wikipedia, specialized topics like COVID-19, different text types like news topics and tweets, and datasets with different sizes, query lengths, and document lengths.

This benchmark is used to measure the performance of our systems and have comparable results. Additionally, the benchmark can be utilized in two distinct manners: zero-shot evaluation, which involves training a system on a specific dataset (typically MS-MARCO) and evaluating its performance using all the BEIR test datasets; and unsupervised domain adaptation, where we use the documents of one of the datasets to generate queries and create training and development splits, then we train an IR model, and finally we evaluate it using the BEIR test dataset. 

In this paper, we will apply unsupervised domain adaptation to all the datasets on the BEIR benchmark independently. In the case of CQADupStack benchmark \cite{cqadupstack}, we train a model for each dataset as the StackExchange community covers a wide variety of topics such as android, english, gaming, programming, and so on.

\subsection{Metrics}
Typical classification and regression metrics measure whether the predicted value is close to the actual value. Unfortunately, these metrics do not take into consideration the order of prediction, which is important in IR, as it is not the same to find the answer in the first document or in the last one. To evaluate an IR system we need to measure how relevant the results are and how good the ordering is.

The most used metric in IR systems, used also in BEIR, is the nDCG@K metric, more specifically nDCG@10 with k=10. \citet{CONTRIEVER} claim that nDCG@K is good at evaluating rankings returned to humans, for example in a search engine, but that Recall@100 is relevant to evaluate retrievers that are used in machine learning systems, such as question answering. These two metrics are important as humans only check the first results provided, meaning that is important the order the documents are returned (nDCG@10 metric). On the other hand, question answering systems do not consider the number or order of the documents passed, making Recall@100 more informative. In this paper, we provide both metrics.

\section{Experiments and Results}

We train three different types of systems, namely Marco-Supervised, Marco-Unsupervised, and Marco-with-Unsupervised-DA. For the unsupervised setting we have two variants, LLM and independent cropping, depending on the method employed for query generation. We follow usual practice and train a supervised version of the system using MS-MARCO dataset, named Marco-Supervised, which will be used as the base system for unsupervised DA. We leverage this supervised version as starting point because for real world applications this dataset is publicly available, and yields strong performance \cite{MSMARCO}. Therefore, we refrain from conducting experiments involving completely unsupervised methodologies, such as marco-unsupervised + Unsupervised Domain Adaptation, or training only on BEIR datasets. 

Two unsupervised DA systems are trained named $DA_{LLM}$ 
for queries generated using Large Language models and $DA_{Cropping}$ for queries generated using the independent cropping generation method. Both of these systems fine-tune the Marco-Supervised system.

For independent cropping the queries are generated using the default parameters found in the official implementation. In the case of LLM we need to evaluate which checkpoint is the best one.

\paragraph{Choosing the best checkpoint for zero-shot query generation using LLM}

\leavevmode\newline

\begin{figure}[htbp] 
  \begin{center}
    \includegraphics[width=8.5cm]{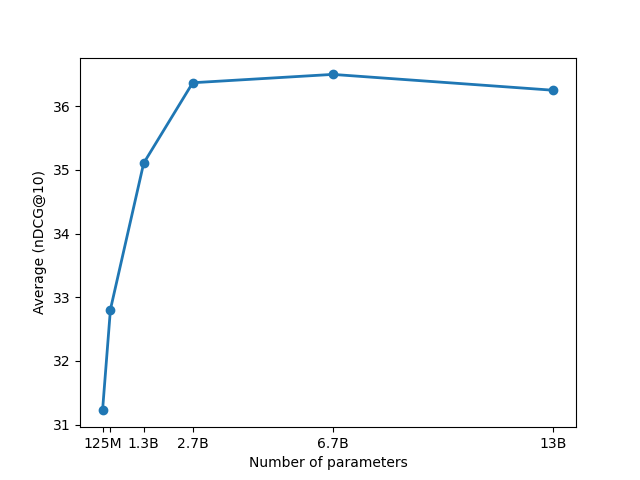} 
    \caption{Results on BEIR of the unsupervised LLM system for the zero-shot scenario. The average performance of OPT LLMs of different sizes are shown.} 
    \label{fig:average_opt} 
  \end{center} 
\end{figure}

In order to choose which OPT checkpoint needs to be used for query generation we perform experiments on all the available checkpoints. Figure \ref{fig:average_opt} compares the average performance of the different OPT checkpoints on the BEIR benchmark in a zero-shot setting.

According to the figure, the average scores for the 2.7B, 6.7B, and 13B checkpoints are similar, with the 6.7B checkpoint performing the best, followed by the 2.7B checkpoint with a difference of only 0.131 points. However, the scores reach a plateau after the 2.7B checkpoint. Considering that (i) the difference between the 2.7B and 6.7B checkpoints is negligible and (ii) the time required to generate the datasets is almost halved when using the 2.7B checkpoint (see Figure \ref{fig:parameters_vs_time}), we decide to use the 2.7B checkpoint for DA. The scores obtained for each dataset can be found in Appendix \ref{a_BEIR_RESULTS_NDCG_COMPARE_OPT}.

\paragraph{Results}

\leavevmode\newline

The results comparing the zero-shot and unsupervised domain adaptation performance can be found in Table \ref{BEIR_RESULTS_NDCG_COMPARE}. The results for the Recall@100 metric can be found on Appendix \ref{recall_results}.

\begin{table*}[htbp]
    \centering
    \begin{tabular}{c|ccc|cc}
    \toprule
    &   \multicolumn{3}{c|}{\bf Zero-shot} &   \multicolumn{2}{c}{\bf Domain Adaptation} \\
    &  \bf Marco- & \multicolumn{2}{c|}{\bf Marco-Unsupervised} &   \multicolumn{2}{c}{\bf Unsupervised DA} \\
    &  \bf Supervised & \bf LLM  & \bf  Cropping & \bf LLM &   \bf Cropping \\
    \midrule
    Wins & 3 & 2 & 0 & \bf 9 & 0 \\
    Average       &  \underline{37.31}         & 36.37       &  18.98     &  \bf 40.47     &  23.53        \\
    \midrule
    TREC-Covid    &  \underline{45.51}          & 41.01 &  15.09         &   \bf 56.75     &  38.27        \\
    NFCorpus      &  27.09          & \bf 32.24 &  21.94         &  \underline{28.34}     &  23.99        \\
    Natural Questions  &  \bf 34.36     & 26.00 &  10.04         &  \underline{27.87}     &  11.99        \\
    HotpotQA      &  \underline{46.33}          & 45.61  &  9.61         &  \bf 48.19     &  22.10        \\
    FiQA          &  22.31          & \underline{24.70} &  7.65          &  \bf 25.93     &  10.45        \\
    ArguAna       &  42.20          & \underline{45.02} & 33.77          &  \bf 53.96     &  41.63        \\
    Tóuche-2020   & \underline{13.87}             & 11.01 & 2.82           &  \bf 21.26     &  2.61         \\
    CQAdupstack   &  \underline{27.62}          & 26.66 & 17.81        &  \bf 32.17     &  18.53        \\
    Quora         &  \underline{82.90}          & 80.37  & 72.00         &  \bf 84.60     &  66.37        \\
    DBpedia       & \underline{30.77}                 & 30.56  & 10.77         &  \bf 32.51     &  14.04        \\
    Scidocs       &  12.97          & \underline{13.51} & 7.32          &  \bf 15.20     &  8.49         \\
    Fever         &  \bf 63.37          & 50.73  & 7.05          &  \underline{60.47}     &  8.75         \\
    Climate-Fever &  \bf 21.64          & 21.33  & 2.83          &  \underline{21.48}     &  10.13        \\
    Scifact       &  51.42          & \bf 60.39  & 47.02         &  \underline{57.80}     &  52.04        \\
    \bottomrule
    \end{tabular}
    \caption{Results for all five systems on each dataset of BEIR for the two scenarios. Three systems (one supervised and two unsupervised) in the zero-shot scenario, and two unsupervised DA systems (DA scenario). Number of wins in each dataset and average performance are also shown. In each row, bold for best system, underline for second best. }
    \label{BEIR_RESULTS_NDCG_COMPARE}
\end{table*}

\paragraph{Overall results for zero-shot performance}

\leavevmode\newline

The strong results of the unsupervised system trained using the queries generated by the LLM are noteworthy, as they are less than a point from those of the supervised system trained on MS-MARCO. This confirms the validity of using queries generated by LLMs. On the other hand the results for the unsupervised system trained cropping are much worse. 

\paragraph{Overall results for unsupervised domain adaptation}

\leavevmode\newline

Regarding unsupervised domain adaptation, the $DA_{LLM}$ method outperforms the Marco-Supervised approach in average by 3.356 points in the nDCG@10 metric. In the case of $DA_{Cropping}$, the system performs worse by -13.782 points, but it is still better than the Marco-Unsupervised-Cropping approach. These experiments demonstrate that unsupervised domain adaptation can enhance performance, especially when the LLM method is used.

To summarize, we conclude that LLM works better than independent cropping, and the difference between the Marco-Supervised and Marco-Unsupervised approaches is negligible. However, when unsupervised domain adaptation is used with a LLM, the performance can be improved by 3 extra points. 

In addition to providing the average metric, we also report the \emph{wins} metric to better evaluate the performance of the system. The average metric may not be the most appropriate in certain cases, as datasets can vary in size and domain, and one dataset with exceptional performance can heavily influence the overall average, even if other datasets perform poorly. Our evaluation reveals that $DA_{LLM}$outperforms the other systems in 9 out of 14 datasets and ranks as the second best in the remaining 5 datasets. In contrast, the systems using independent cropping fail to achieve a win in any of the datasets.

\section{Analysis of query generation}

In this section, we analyze the queries generated by both independent cropping and LLM.

\subsection{Questions lost during generation}
\begin{figure}[htbp] 
  \begin{center}
    \includegraphics[width=8cm]{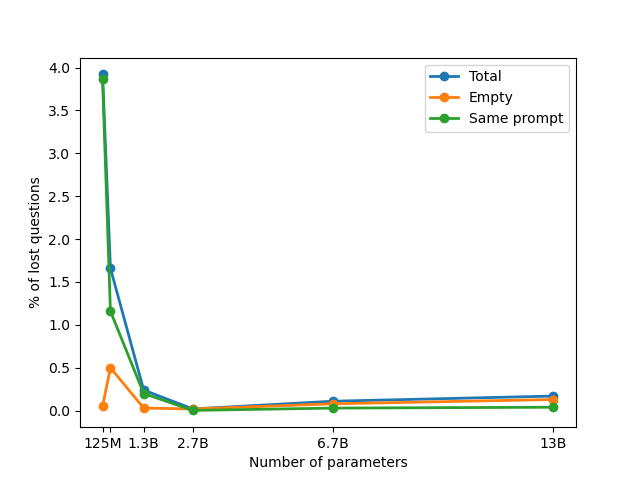} 
    \caption{The percentage of number of questions lost during inference from a corpus of 8,841,661 documents (MS-MARCO).} 
    \label{fig:opt-generated} 
  \end{center} 
\end{figure}

Figure \ref{fig:opt-generated} shows the queries lost during generation for the MS-MARCO dataset. We postprocess the dataset taking into account empty queries (the system is not able to generate a query) and same prompt queries (the system generates the same query provided in one of the examples of the prompt).

In the figure, we can see that the number of lost questions decreases as the number of parameters increases. Nevertheless, from the 2.7B checkpoint onwards, we can see that the number of lost questions increases, mainly because no query is generated, but also because the same query is generated as one of the examples of the prompt.

In the case of independent cropping, we see that a small amount of queries (0.02\%) are lost probably due to short documents being used and the random deletions of words deleting the remaining ones.

\subsection{Top 10 starting words}

In English, wh- words are used the most for question formation. This is because they are interrogative pronouns or adverbs that are used to ask for information. The wh- words include what, who, whom, whose, which, when, where, why and how. They are versatile and can be used to ask about a wide range of information, such as the identity of a person, the location of an event, the reason for an action, or the manner in which something is done. As a result, wh- words are essential for forming questions and seeking knowledge in the English language.

In Table \ref{top_10_starting_words}, we can see the top 10 starting words for all the OPT checkpoints. Overall, we can see that the same starting words are used in all the checkpoints, with word "what" being used the most and the frequency of it increasing with the number of parameters. 

\begin{table*}[htbp]
    \hspace{-2cm}
        \begin{tabular}{ccccccc}
        \toprule
        \bf Top 10 words &   \bf 125M &   \bf 350M &   \bf 1.3B &   \bf 2.7B &   \bf 6.7B &   \bf 13B \\
        \midrule
        Top 1   &   What (33.5\%)   &   What (36.93\%)   &   What (38.35\%)   &   What (48.73\%)    &   What (52.7\%)   &   What (56.68\%)  \\
        Top 2   &   How (18.03\%)   &   How (31.72\%)    &   How (25.49\%)    &   How   (18.41\%)   &   How (19.66\%)   &   How (20.37\%)   \\
        Top 3   &   Is (9.48\%)     &   Why (5.38\%)     &   Is  (6.46\%)     &   Is    (5.82\%)    &   Which (4.14\%)  &   Who (3.39\%)    \\
        Top 4   &   Why (5.62\%)    &   Is (4.33\%)      &   Can (4.68\%)     &   Why   (3.46\%)    &   Is (4.05\%)     &   Is (3.13\%)     \\
        Top 5   &   Can (5.37\%)    &   The (2.54\%)     &   Why (4.53\%)     &   Who   (2.92\%)    &   Who (3\%)       &   When (2.43\%)   \\
        Top 6   &   Do (4.88\%)     &   Does (2.36\%)    &   Which (2.83\%)   &   Which (2.88\%)    &   Where (2.9\%)   &   Where (2.34\%)  \\
        Top 7   &   Does (4.67\%)   &   Can (2.31\%)     &   Does (2.61\%)    &   What's (2.56\%)   &   What's (2.23\%) &   Which (2.07\%)  \\
        Top 8   &   Are (4.03\%)    &   Are (1.71\%)     &   What's (2.19\%)  &   Can  (2.39\%)     &   Can (1.23\%)    &   Can (1.64\%)    \\
        Top 9   &   Who (2.54\%)    &   Do (1.69\%)      &   Where (2.13\%)   &   Where (2.38\%)    &   When (0.98\%)   &   What's (1.62\%) \\
        Top 10  &   If (2.47\%)     &   Who (1.49\%)     &   Do (1.93\%)      &   When (1.15\%)     &   Why (0.93\%)    &   Why (1.19\%)    \\
        \midrule
        Total   &   90.59\%         &   90.46\%          &   91.2\%           &   90.7\%            &   91.82\%         &   94.86\%         \\
        \bottomrule
        \end{tabular}
    
    \caption{\label{top_10_starting_words} Top 10 starting words for each OPT checkpoint.}
\end{table*}

\subsection{Question generation time}

The time required to generate a question using large language models (LLMs) typically increases with the amount of parameters in the model. This is because LLMs with more parameters require more computational power to generate each query, and therefore take longer to generate questions. This can be a bottleneck for real-world applications that require fast and efficient question generation, especially when dealing with large datasets or when generating a large number of queries. Thus, finding ways to balance the trade-off between model size and computational efficiency is an important consideration when using LLMs for question generation.

\begin{figure}[t]
  \begin{center}
    \includegraphics[width=8cm]{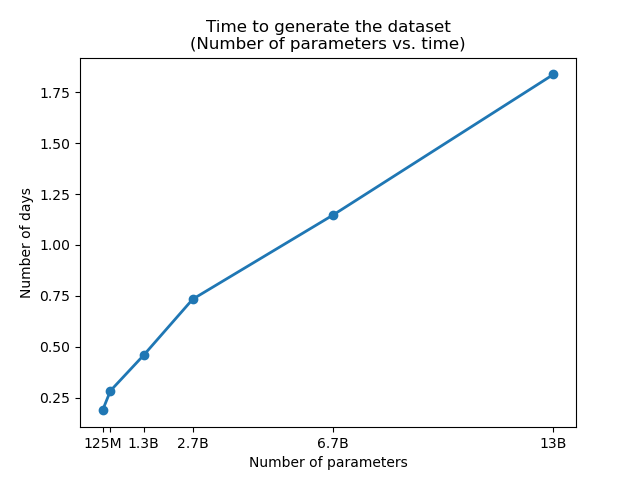} 
    \caption{Graph showing the time required for generating the dataset using different OPT checkpoints. To generate the dataset, we are making use of 8 Nvidia A100 GPUs with 80GB of VRAM memory each.} 
    \label{fig:parameters_vs_time} 
  \end{center} 
\end{figure}

In this section, we measure the time required to generate questions using all the OPT checkpoints from 125 million to 13 billion parameters. Figure \ref{fig:parameters_vs_time} shows the time required to generate each dataset using 8 Nvidia A100 GPUs with 80 GB of VRAM.

\section{Conclusions and Future Work}

In this paper, we have explored the use of Large Language Models (LLMs) for synthetic query generation and compared them to the rule-based independent cropping method for unsupervised domain adaptation. Our results show that LLM-based methods outperform independent cropping in all scenarios by a significant margin. Although LLMs require more time, the benefits they offer in terms of performance make them a viable alternative for query generation.

We also studied the impact of the number of parameters on the performance of our system when a LLM is used for generating queries. Our experiments indicate that beyond a certain point, increasing the number of parameters in our models did not lead to any significant increase in performance. Specifically, we found that the performance of our models did not improve beyond 2.7B parameters.

Furthermore, we conducted our experiments using open-source LLMs, which allowed us to compare our results with those of other researchers and ensure reproducibility.

Finally, we demonstrated that unsupervised domain adaptation is an effective approach for improving the performance of neural IR systems, as compared to zero-shot learning. Our experiments showed that fine-tuning a supervised system in synthetic data from the target domain leads to significant performance improvements.

For future work, we plan to explore the use of other sampling techniques and strategies for query generation using LLMs. Additionally, we will investigate different query generation techniques to further improve our system's performance.

We also plan to study the impact of other types of sentence similarities on the performance of our models. Furthermore, we will examine the use of retrievers with more parameters and the use of rerankers to improve performance.

\section*{Acknowledgements}

Carlos is funded by a PhD grant from the Spanish Government (FPU21/02867). This work is partially supported by the Ministry of Science and Innovation of the Spanish Government (AWARE project TED2021-131617B-I00, DeepKnowledge project PID2021-127777OB-C21), and the Basque Government (IXA excellence research group IT1570-22).

\bibliography{acl_latex}

\begin{thebibliography}{29}
\expandafter\ifx\csname natexlab\endcsname\relax\def\natexlab#1{#1}\fi

\bibitem[{Black et~al.(2021)Black, Leo, Wang, Leahy, and Biderman}]{gpt_neo}
Sid Black, Gao Leo, Phil Wang, Connor Leahy, and Stella Biderman. 2021.
\newblock {GPT-Neo: Large Scale Autoregressive Language Modeling with Mesh-Tensorflow}.

\bibitem[{Black et~al.(2022)Black, Biderman, Hallahan, Anthony, Gao, Golding, He, Leahy, McDonell, Phang et~al.}]{gpt_neox}
Sidney Black, Stella Biderman, Eric Hallahan, Quentin Anthony, Leo Gao, Laurence Golding, Horace He, Connor Leahy, Kyle McDonell, Jason Phang, et~al. 2022.
\newblock Gpt-neox-20b: An open-source autoregressive language model.
\newblock In \emph{Proceedings of BigScience Episode$\backslash$\# 5--Workshop on Challenges \& Perspectives in Creating Large Language Models}, pages 95--136.

\bibitem[{Bonifacio et~al.(2022)Bonifacio, Abonizio, Fadaee, and Nogueira}]{INPARS}
Luiz Bonifacio, Hugo Abonizio, Marzieh Fadaee, and Rodrigo Nogueira. 2022.
\newblock Inpars: Data augmentation for information retrieval using large language models.

\bibitem[{Brown et~al.(2020)Brown, Mann, Ryder, Subbiah, Kaplan, Dhariwal, Neelakantan, Shyam, Sastry, Askell et~al.}]{GPT3}
Tom Brown, Benjamin Mann, Nick Ryder, Melanie Subbiah, Jared~D Kaplan, Prafulla Dhariwal, Arvind Neelakantan, Pranav Shyam, Girish Sastry, Amanda Askell, et~al. 2020.
\newblock Language models are few-shot learners.
\newblock \emph{Advances in neural information processing systems}, 33:1877--1901.

\bibitem[{Choi et~al.(2021)Choi, Kim, Joe, and Gwon}]{cls_bad}
Hyunjin Choi, Judong Kim, Seongho Joe, and Youngjune Gwon. 2021.
\newblock Evaluation of bert and albert sentence embedding performance on downstream nlp tasks.
\newblock In \emph{2020 25th International conference on pattern recognition (ICPR)}, pages 5482--5487. IEEE.

\bibitem[{Dai et~al.(2023)Dai, Zhao, Ma, Luan, Ni, Lu, Bakalov, Guu, Hall, and Chang}]{promptagator}
Zhuyun Dai, Vincent~Y Zhao, Ji~Ma, Yi~Luan, Jianmo Ni, Jing Lu, Anton Bakalov, Kelvin Guu, Keith Hall, and Ming-Wei Chang. 2023.
\newblock Promptagator: Few-shot dense retrieval from 8 examples.
\newblock In \emph{The Eleventh International Conference on Learning Representations}.

\bibitem[{Devlin et~al.(2019)Devlin, Chang, Lee, and Toutanova}]{bert}
Jacob Devlin, Ming-Wei Chang, Kenton Lee, and Kristina Toutanova. 2019.
\newblock {BERT}: Pre-training of deep bidirectional transformers for language understanding.
\newblock In \emph{Proceedings of the 2019 Conference of the North {A}merican Chapter of the Association for Computational Linguistics: Human Language Technologies, Volume 1 (Long and Short Papers)}, pages 4171--4186, Minneapolis, Minnesota. Association for Computational Linguistics.

\bibitem[{Formal et~al.(2021)Formal, Piwowarski, and Clinchant}]{splade}
Thibault Formal, Benjamin Piwowarski, and St{\'e}phane Clinchant. 2021.
\newblock Splade: Sparse lexical and expansion model for first stage ranking.
\newblock In \emph{Proceedings of the 44th International ACM SIGIR Conference on Research and Development in Information Retrieval}, pages 2288--2292.

\bibitem[{Furnas et~al.(1987)Furnas, Landauer, Gomez, and Dumais}]{vocabulary_mismatch}
George~W. Furnas, Thomas~K. Landauer, Louis~M. Gomez, and Susan~T. Dumais. 1987.
\newblock The vocabulary problem in human-system communication.
\newblock \emph{Communications of the ACM}, 30(11):964--971.

\bibitem[{Gulzar et~al.(2018)Gulzar, Leema, and Deepak}]{recommender_systems}
Zameer Gulzar, A~Anny Leema, and Gerard Deepak. 2018.
\newblock Pcrs: Personalized course recommender system based on hybrid approach.
\newblock \emph{Procedia Computer Science}, 125:518--524.

\bibitem[{Hoogeveen et~al.(2015)Hoogeveen, Verspoor, and Baldwin}]{cqadupstack}
Doris Hoogeveen, Karin~M. Verspoor, and Timothy Baldwin. 2015.
\newblock Cqadupstack: A benchmark data set for community question-answering research.
\newblock In \emph{Proceedings of the 20th Australasian Document Computing Symposium}, ADCS '15, New York, NY, USA. Association for Computing Machinery.

\bibitem[{Izacard et~al.(2022)Izacard, Caron, Hosseini, Riedel, Bojanowski, Joulin, and Grave}]{CONTRIEVER}
Gautier Izacard, Mathilde Caron, Lucas Hosseini, Sebastian Riedel, Piotr Bojanowski, Armand Joulin, and Edouard Grave. 2022.
\newblock Unsupervised dense information retrieval with contrastive learning.
\newblock \emph{Transactions on Machine Learning Research}.

\bibitem[{Izacard and Grave(2021)}]{FiD}
Gautier Izacard and {\'E}douard Grave. 2021.
\newblock Leveraging passage retrieval with generative models for open domain question answering.
\newblock In \emph{Proceedings of the 16th Conference of the European Chapter of the Association for Computational Linguistics: Main Volume}, pages 874--880.

\bibitem[{Karpukhin et~al.(2020)Karpukhin, Oguz, Min, Lewis, Wu, Edunov, Chen, and Yih}]{DPR}
Vladimir Karpukhin, Barlas Oguz, Sewon Min, Patrick Lewis, Ledell Wu, Sergey Edunov, Danqi Chen, and Wen-tau Yih. 2020.
\newblock Dense passage retrieval for open-domain question answering.
\newblock In \emph{Proceedings of the 2020 Conference on Empirical Methods in Natural Language Processing (EMNLP)}, pages 6769--6781.

\bibitem[{Khattab and Zaharia(2020)}]{colbert}
Omar Khattab and Matei Zaharia. 2020.
\newblock Colbert: Efficient and effective passage search via contextualized late interaction over bert.
\newblock In \emph{Proceedings of the 43rd International ACM SIGIR conference on research and development in Information Retrieval}, pages 39--48.

\bibitem[{Kobayashi and Takeda(2000)}]{ir_for_search_engines}
Mei Kobayashi and Koichi Takeda. 2000.
\newblock Information retrieval on the web.
\newblock \emph{ACM computing surveys (CSUR)}, 32(2):144--173.

\bibitem[{Lee et~al.(2019)Lee, Chang, and Toutanova}]{ICT}
Kenton Lee, Ming-Wei Chang, and Kristina Toutanova. 2019.
\newblock Latent retrieval for weakly supervised open domain question answering.
\newblock In \emph{Proceedings of the 57th Annual Meeting of the Association for Computational Linguistics}, pages 6086--6096.

\bibitem[{Nguyen et~al.(2016)Nguyen, Rosenberg, Song, Gao, Tiwary, Majumder, and Deng}]{MSMARCO}
Tri Nguyen, Mir Rosenberg, Xia Song, Jianfeng Gao, Saurabh Tiwary, Rangan Majumder, and Li~Deng. 2016.
\newblock {MS} {MARCO:} {A} human generated machine reading comprehension dataset.
\newblock \emph{CoRR}, abs/1611.09268.

\bibitem[{Nogueira et~al.(2020)Nogueira, Jiang, Pradeep, and Lin}]{monoT5}
Rodrigo Nogueira, Zhiying Jiang, Ronak Pradeep, and Jimmy Lin. 2020.
\newblock Document ranking with a pretrained sequence-to-sequence model.
\newblock In \emph{Findings of the Association for Computational Linguistics: EMNLP 2020}, pages 708--718.

\bibitem[{Raffel et~al.(2020)Raffel, Shazeer, Roberts, Lee, Narang, Matena, Zhou, Li, and Liu}]{T5}
Colin Raffel, Noam Shazeer, Adam Roberts, Katherine Lee, Sharan Narang, Michael Matena, Yanqi Zhou, Wei Li, and Peter~J. Liu. 2020.
\newblock Exploring the limits of transfer learning with a unified text-to-text transformer.
\newblock \emph{Journal of Machine Learning Research}, 21(140):1--67.

\bibitem[{Reimers and Gurevych(2019)}]{sbert}
Nils Reimers and Iryna Gurevych. 2019.
\newblock Sentence-bert: Sentence embeddings using siamese bert-networks.
\newblock In \emph{Proceedings of the 2019 Conference on Empirical Methods in Natural Language Processing and the 9th International Joint Conference on Natural Language Processing (EMNLP-IJCNLP)}, pages 3982--3992.

\bibitem[{Robertson and Zaragoza(2009)}]{BM25}
Stephen Robertson and Hugo Zaragoza. 2009.
\newblock The probabilistic relevance framework: Bm25 and beyond.
\newblock \emph{Foundations and Trends in Information Retrieval}, 3:333--389.

\bibitem[{Sanh et~al.(2019)Sanh, Debut, Chaumond, and Wolf}]{distilbert}
Victor Sanh, Lysandre Debut, Julien Chaumond, and Thomas Wolf. 2019.
\newblock Distilbert, a distilled version of {BERT:} smaller, faster, cheaper and lighter.
\newblock \emph{CoRR}, abs/1910.01108.

\bibitem[{Scao et~al.(2023)Scao, Fan, Akiki, Pavlick, Ilić, Hesslow, Castagné, Luccioni, Yvon, Gallé, Tow, Rush, Biderman, Webson, Ammanamanchi, Wang, Sagot, Muennighoff, Moral, and Wolf}]{BLOOM}
Teven Scao, Angela Fan, Christopher Akiki, Ellie Pavlick, Suzana Ilić, Daniel Hesslow, Roman Castagné, Alexandra Luccioni, François Yvon, Matthias Gallé, Jonathan Tow, Alexander Rush, Stella Biderman, Albert Webson, Pawan Ammanamanchi, Thomas Wang, Benoît Sagot, Niklas Muennighoff, Albert Moral, and Thomas Wolf. 2023.
\newblock Bloom: A 176b-parameter open-access multilingual language model.

\bibitem[{Thakur et~al.(2021)Thakur, Reimers, R{\"u}ckl{\'e}, Srivastava, and Gurevych}]{BEIR}
Nandan Thakur, Nils Reimers, Andreas R{\"u}ckl{\'e}, Abhishek Srivastava, and Iryna Gurevych. 2021.
\newblock Beir: A heterogeneous benchmark for zero-shot evaluation of information retrieval models.
\newblock In \emph{Thirty-fifth Conference on Neural Information Processing Systems Datasets and Benchmarks Track (Round 2)}.

\bibitem[{Wu et~al.(2022)Wu, Luan, Rashkin, Reitter, Hajishirzi, Ostendorf, and Tomar}]{conversational_search}
Zeqiu Wu, Yi~Luan, Hannah Rashkin, David Reitter, Hannaneh Hajishirzi, Mari Ostendorf, and Gaurav~Singh Tomar. 2022.
\newblock {CONQRR}: Conversational query rewriting for retrieval with reinforcement learning.
\newblock In \emph{Proceedings of the 2022 Conference on Empirical Methods in Natural Language Processing}, pages 10000--10014, Abu Dhabi, United Arab Emirates. Association for Computational Linguistics.

\bibitem[{Xu et~al.(2022)Xu, Guo, Duan, and McAuley}]{laprador}
Canwen Xu, Daya Guo, Nan Duan, and Julian McAuley. 2022.
\newblock Laprador: Unsupervised pretrained dense retriever for zero-shot text retrieval.
\newblock In \emph{Findings of the Association for Computational Linguistics: ACL 2022}, pages 3557--3569.

\bibitem[{Yan et~al.(2016)Yan, Duan, Bao, Chen, Zhou, Li, and Zhou}]{chatbots}
Zhao Yan, Nan Duan, Junwei Bao, Peng Chen, Ming Zhou, Zhoujun Li, and Jianshe Zhou. 2016.
\newblock Docchat: An information retrieval approach for chatbot engines using unstructured documents.
\newblock In \emph{Proceedings of the 54th Annual Meeting of the Association for Computational Linguistics (Volume 1: Long Papers)}, pages 516--525.

\bibitem[{Zhang et~al.(2022)Zhang, Roller, Goyal, Artetxe, Chen, Chen, Dewan, Diab, Li, Lin, Mihaylov, Ott, Shleifer, Shuster, Simig, Koura, Sridhar, Wang, and Zettlemoyer}]{OPT}
Susan Zhang, Stephen Roller, Naman Goyal, Mikel Artetxe, Moya Chen, Shuohui Chen, Christopher Dewan, Mona Diab, Xian Li, Xi~Victoria Lin, Todor Mihaylov, Myle Ott, Sam Shleifer, Kurt Shuster, Daniel Simig, Punit~Singh Koura, Anjali Sridhar, Tianlu Wang, and Luke Zettlemoyer. 2022.
\newblock Opt: Open pre-trained transformer language models.

\end{thebibliography}

\appendix

\section{Query generation using Large Language Models}
\label{LLM_extra_info}
\subsection{Using sampling for question generation}
Recent advances in Natural Language Processing (NLP) have led to the development of large language models that can generate human-like text. These models have been successfully applied to various NLP tasks, including text generation, question answering, and IR. However, generating high-quality queries for IR using large language models remains a challenging problem.

One common approach to query generation is to use the argmax function to select the most likely word from the distribution generated by the language model. However, this approach has limitations, as it tends to produce repetitive or generic queries, which may not capture the nuances of the information need.

An alternative approach is to use sampling to generate queries. This approach involves randomly sampling from the distribution generated by the language model to generate a diverse set of queries. Sampling can produce a wider range of queries than argmax, which can improve the chances of finding relevant documents.

In this paper, to include diversity in the questions generated, we apply sampling and a top $p$ of 0.9 (only the smallest set of most probable tokens with probabilities that add up to top $p$ or higher are kept for generation.)

\subsection{Large Language Models}
In this paper, we check the suitability of 4 LLMs: OPT \cite{OPT}, Bloom \cite{BLOOM}, GPT-neo \cite{gpt_neo}, and GPT-neoX \cite{gpt_neox}. In order to evaluate which model is the most appropriate one, one of the things that we need evaluate is how many checkpoints of different parameters each system offers.

\begin{itemize}
    \item \textbf{OPT}: Checkpoints of 125 million, 350 million, 1.3 billion, 2.7 billion, 6.7 billion, 13 billion, 30 billion, 60 billion, and 175 billion parameters.
    \item \textbf{Bloom}: Checkpoints of 560 million, 1.1 billion , 1.77 billion, 3 billion, 7.1 billion, and 176 billion parameters.
    \item \textbf{GPT-neo(X)}: Checkpoints of 125 million, 1.3 billion, 2.7 billion, and 20 billion parameters. We are going to consider GPT-neo and GPT-neoX to be of the same family of models.
\end{itemize}

Looking at the checkpoints available for each transformer, we believe that OPT offers a wider variety of number of parameters.

Next, we compare how each model is able to generate questions. In order to generate questions in an unsupervised way (zero-shot question generation), we need to make use of a prompt as mentioned in \ref{LLM_for_query_generation}. In Table \ref{question_generation}, we compare the capability of different models of similar size in generating questions in an unsupervised setting.

\begin{table*}[htbp]
        \scriptsize
        \begin{tabular}{p{4cm}p{1.2cm}p{1.2cm}p{1.4cm}p{1.4cm}p{1.8cm}p{1.4cm}}
        \toprule
        \bf Document &   \bf OPT-1.3B &   \bf OPT-30B &   \bf Bloom 1.7B &   \bf Bloom 7B &   \bf GPT-neo 1.3B &   \bf GPT-neoX 20B\\
        \midrule
        Color hex is a easy to use tool to get the color codes information including color models (RGB,HSL,HSV and CMYK), css and html color codes. &   How can color hex code help me? &   What color hex is? &   What color is in Hex color model number 0c00a00? &   Html Color Code &   I need to convert the color hex codes to RGB and then I need to do this for all my website color codes for a custom website I am developing. Ar &   What is CSS and HSL? \\
        \midrule
        Although the European powers did make military interventions in Latin America from time to time after the Monroe Doctrine was announced, the Americans did not look for war. They did, however, use the doctrine as justification for taking Texas in 1842 under President John Tyler. &   What is the Monroe Doctrine? &   How was the Monroe Doctrine used in the 1840s? &   When did the US and European powers make military intervention in Latin America? &   What country supported a campaign that gave Texas into Mexico's hands? &   What is the Monroe Doctrine? &   When did the United States make military interventions in Latin America? \\
        \bottomrule
        \end{tabular}
    
    \caption{\label{question_generation} Comparison of automatically generated questions using different models in a zero-shot setting.}
\end{table*}

Overall, in a limited qualitative analysis, we can see that the quality of generation of OPT is better than Bloom and GPT-neo(X). The reasons are the following: first, Bloom 7B and GPT-neo 1.3B are not able to generate a question for the first document, second, the question generated by Bloom 7B in the third document is not correct, and finally, OPT-30B seems to be generating more diverse questions as it is using different starting words compared to Bloom and GPT-neo(X) \footnote{This reasoning comes from a bigger sample of questions}. Given this initial analysis, and mainly because of the variety of checkpoints available we decided to focus on OPT as our query generation LLM.

\subsection{Best OPT checkpoint}
\label{a_BEIR_RESULTS_NDCG_COMPARE_OPT}
Table \ref{BEIR_RESULTS_NDCG_COMPARE_OPT} shows the performance of the different OPT checkpoints on the BEIR benchmark using an out-of-domain methodology. In the table we observe that that the 2.7B, 6.7B, and 13B perform really similar on the average. Nevertheless, the 2.7B checkpoint wins in most of the datasets.

\begin{table*}[htbp]
    \centering
    \begin{tabular}{c|cccccc}
    \toprule
    \bf Dataset \symbol{92} Model Name &   \bf OPT-125M &   \bf OPT-350M &   \bf OPT-1.3B &   \bf OPT-2.7B &   \bf OPT-6.7B &   \bf OPT-13B \\
    \midrule
    MS-MARCO &   21.17 &   22.97 &   23.72 &   \underline{25.70} &   \textbf{25.88} &   24.58 \\
    \midrule
    Wins & 1 & 0 & 1 & \bf 5 & 4 & 3 \\
    \midrule
    Average &   31.22 &   32.81 &   35.11 &   \underline{36.37} &   \textbf{36.50} &   36.25 \\
    \midrule
    TREC-Covid &   36.26 &   36.79 &   37.90 &   \textbf{41.01} &   \underline{38.43} &   35.41 \\
    NFCorpus &   28.85 &   28.73 &   32.02 &   32.24 &   \textbf{33.23} &   \underline{33.02} \\
    Natural Questions &   18.49 &   20.89 &   23.31 &   26.00 &   \textbf{27.15} &   \underline{26.64} \\
    HotpotQA &   36.58 &   39.32 &   42.73 &   \textbf{45.61} &   \underline{45.28} &   45.19 \\
    FiQA &   21.20 &   21.15 &   24.92 &   24.70 &   \underline{26.11} &   \textbf{26.15} \\
    ArguAna &   45.41 &   \underline{47.80} &   \textbf{48.57} &   45.02 &   47.05 &   47.15 \\
    Tóuche-2020 &   \textbf{12.19} &   10.97 &   10.57 &   \underline{11.01} &   10.18 &   10.53 \\
    CQAdupstack &   22.35 &   24.34 &   25.71 &   26.66 &   \textbf{27.49} &   \underline{27.40} \\
    Quora &   74.62 &   76.59 &   79.46 &   80.37 &   \underline{81.29} &   \textbf{81.65} \\
    DBpedia &   25.41 &   27.46 &   29.56 &   \textbf{30.56} &   29.72 &   \underline{30.32} \\
    Scidocs &   13.12 &   12.65 &   13.40 &   13.51 &   \underline{13.59} &   \textbf{13.75} \\
    Fever &   33.13 &   43.11 &   46.83 &   50.73 &   \textbf{53.10} &   \underline{51.12} \\
    Climate-Fever &   14.81 &   15.45 &   18.91 &   \textbf{21.33} &   19.79 &   \underline{20.43} \\
    Scifact &   54.71 &   54.04 &   57.58 &   \textbf{60.39} &   58.56 &   \underline{58.71} \\
    \bottomrule
    \end{tabular}
    \caption{\label{BEIR_RESULTS_NDCG_COMPARE_OPT} Table comparing the performance of different OPT checkpoints on BEIR benchmark. All scores denote nDCG@10. The best score is marked in bold and the second best is underlined. The average corresponds to the average of all the datasets except MS MARCO.}
\end{table*}

\section{Models: IR model}
\label{model_information}

In this section, we provide information about the models that were trained in this paper.
\subsection{DPR vs. SBERT}
\label{dpr_vs_sbert}
Dense passage retrieval (DPR) and Sentence-BERT (SBERT) are both methods for generating representations of text that can be used in NLP applications such as IR, question-answering, and text classification. However, they differ in their approach and the type of representations they generate.

DPR is a method that generates dense representations of long passages of text, typically entire articles or documents. It is designed to efficiently search large collections of text and retrieve relevant passages in response to user queries.

On the other hand, SBERT is a method for generating dense representations of individual sentences. It is designed to capture the meaning of a sentence in a high-dimensional vector space that can be used for various downstream tasks

Both systems use a siamese neural network architecture that takes a pair of sentences as input and outputs a similarity score between them. The network is trained on large-scale datasets using various techniques such as triplet loss and contrastive loss to ensure that similar sentences are mapped to nearby points in the vector space.

The main difference between these two is that DPR trains two encoders, one for the query and another one for the passage, whereas SBERT trains only one encoder that is used for both the queries and passages (this encoder is used independently twice, once for the query and once for passage, making it a bi-encoder model with shared weights). Table \ref{BEIR_dpr_sbert} compares the performance between using DPR and SBERT with MS-MARCO supervised dataset and cosine similarity using zero-shot evaluation.

\begin{table}[htbp]
    \centering
    \begin{tabular}{c|cccc}
    \toprule
    \bf Dataset \symbol{92} Model Name &   \bf DPR &   \bf SBERT \\
    \midrule
    MS-MARCO &   \bf 31.64 &   31.38 \\
    \midrule
    Wins & 5 & \bf 9\\ 
    \midrule
    Average &   \bf 38.03 &   37.31 \\
    \midrule
    TREC-covid &   \bf 62.99 &   45.51 \\
    NFCorpus &   26.72 &   \bf 27.09 \\
    Natural Questions &   \bf 36.01 &   34.36 \\
    HotpotQA &   44.86 &   \bf 46.33 \\
    FiQA &   21.68 &   \bf 22.31 \\
    ArguAna &   38.63 &   \bf 42.20 \\
    Tóuche-2020 &   \bf 18.78 &   13.87 \\
    CQAdupstack &   26.89 &   \bf 27.62 \\
    Quora &   78.01 &   \bf 82.90 \\
    DBpedia &   \bf 31.23 &   30.77 \\
    Scidocs &   12.71 &   \bf 12.97 \\
    Fever &   63.23 &   \bf 63.37 \\
    Climate-fever &   18.20 &   \bf 21.64 \\
    Scifact &   \bf 52.44 &   51.42 \\
    \bottomrule
    \end{tabular}
    \caption{\label{BEIR_dpr_sbert} Table comparing the performance of using DPR and SBERT architectures on the BEIR benchmark. All scores denote NDCG@10. The best score is marked in bold. The average corresponds to the average of all the datasets except MS MARCO. The system used to train and compare both systems is the marco-supervised system.}
\end{table}

The table shows that the difference between the average performance of the two systems is negligible, with only 0.717 points of difference. This extra points come from the TREC-Covid dataset, where we can see that DPR obtains 17.478 point improvement. Therefore, based on the overall performance across 14 datasets, we decide to utilize SBERT, given its superior performance in 9 out of the 14 datasets.

\subsection{Similarity}
Cosine similarity and dot product are two commonly used metrics to measure the similarity between two vectors in machine learning and IR.

Cosine similarity is a measure of similarity between two non-zero vectors of an inner product space that measures the cosine of the angle between them. It is defined as the dot product of the two vectors divided by the product of their magnitudes, as shown in Equation \ref{cosine_similarity}.

\begin{equation}
\label{cosine_similarity}
cos\_similarity(\mathbf{A}, \mathbf{B}) = \frac{\mathbf{A} \cdot \mathbf{B}}{|\mathbf{A}| |\mathbf{B}|}
\end{equation}

The cosine similarity measure ranges from -1 to 1, where a value of 1 means the vectors are identical, 0 means they are orthogonal (i.e., have no correlation), and -1 means they are completely dissimilar.

Dot product is another measure of similarity between two vectors, and it is defined as the sum of the element-wise products of the two vectors, as shown in Equation \ref{dot_score}.

\begin{equation}
\label{dot_score}
dot\_similarity(\mathbf{A}, \mathbf{B}) = \sum_{i=1}^n a_i b_i
\end{equation}

Like cosine similarity, the dot product measures the degree of similarity between the two vectors, but it does not normalize the vectors based on their magnitudes, and thus can be affected by the scale of the vectors.

Table \ref{BEIR_RESULTS_NDCG_COMPARE_SIMILARITY} compares which method is better using MS-MARCO supervised and SBERT model.

\begin{table*}[htbp]
    \centering
    \begin{tabular}{c|ccc|ccc}
    \toprule
    &   \multicolumn{3}{c|}{\bf Dot product similarity} &   \multicolumn{3}{c}{\bf Cosine similarity} \\
    \bf Dataset \symbol{92} Model Name &   \bf supervised &   \bf Cropping &  \bf OPT-2.7B &   \bf supervised &   \bf Cropping &  \bf OPT-2.7B \\
    \midrule
    MS-MARCO & 30.19 & 10.09 & 24.02 & \textbf{31.38} & \textbf{10.95} & \textbf{25.70} \\
    \midrule
    Average & \textbf{37.51} & 14.63 & 32.34 & 37.31 & \textbf{18.98} & \textbf{36.37} \\
    \midrule
    TREC-Covid & \textbf{53.34} & \textbf{20.70} & 26.66 & 45.51 & 15.08 & \textbf{41.01} \\
    NFCorpus & 26.98 & 7.40 & 30.83 & \textbf{27.09} & \textbf{21.94} & \textbf{32.24} \\
    Natural Questions & 32.61 & \textbf{10.11} & 20.59 & \textbf{34.36} & 10.04 & \textbf{26.00} \\
    HotpotQA & \textbf{48.90} & \textbf{16.17} & \textbf{46.37} & 46.33 & 9.61 & 45.61 \\
    FiQA & 22.14 & 6.36 & 25.17 & \textbf{22.31} & \textbf{7.65} & \textbf{24.70} \\
    ArguAna & 39.34 & 1.20 & 38.27 & \textbf{42.20} & \textbf{33.77} & \textbf{45.02} \\
    Tóuche-2020 & \textbf{16.89} & 2.23 & 6.05 & 13.87 & \textbf{2.82} & \textbf{11.01} \\
    CQAdupstack & \textbf{28.82} & \textbf{19.71} & \textbf{27.28} & 27.62 & 17.81 & 26.66 \\
    Quora & \textbf{83.03} & 58.40 & 80.14 & 82.90 & \textbf{72.00} & \textbf{80.37} \\
    DBpedia & 28.52 & 8.59 & 24.61 & \textbf{30.77} & \textbf{10.77} & \textbf{30.56} \\
    Scidocs & \textbf{13.12} & 5.95 & 11.88 & 12.97 & \textbf{7.32} & \textbf{13.51} \\
    Fever & 62.22 & \textbf{11.40} & 43.21 & \textbf{63.37} & 7.054 & \textbf{50.73} \\
    Climate-Fever & 15.77 & \textbf{5.50} & 13.10 & \textbf{21.64} & 2.83 & \textbf{21.33} \\
    Scifact & \textbf{53.48} & 31.06 & 58.60 & 51.42 & \textbf{47.02} & \textbf{60.39} \\
    \bottomrule
    \end{tabular}
    \caption{\label{BEIR_RESULTS_NDCG_COMPARE_SIMILARITY} Table comparing the performance of cosine and dot product similarity on different OPT checkpoint on BEIR benchmark in a zero-shot setting. All scores denote nDCG@10. The best score for each method is marked in bold. The average corresponds to the average of all the datasets except MS MARCO.}
\end{table*}

In the table, we can see that the difference between the two similarities is minimum for the supervised setting. However, when an unsupervised method for query generation is employed, the cosine similarity outperforms the alternative, yielding an improvement of 4.353 and 4.027 points for cropping and OPT-2.7B, respectively. Comparable trends were also observed for other OPT checkpoints.

\section{Results}

\subsection{Recall@100}
\label{recall_results}

Table \ref{BEIR_RESULTS_RECALL_COMPARE} shows the Recall@100 metric of the trained models.

\begin{table*}[htbp]
    \centering
    \begin{tabular}{c|ccc|cc}
    \toprule
    &   \multicolumn{3}{c|}{\bf Zero-shot} &   & \\
    &  \bf Marco- & \multicolumn{2}{c|}{\bf Marco-Unsupervised} &   \multicolumn{2}{c}{\bf Unsupervised DA} \\
    &  \bf Supervised & \bf LLM  & \bf  Cropping & \bf LLM &   \bf Cropping \\
    \midrule
    Wins & 1 & 2 & 0 & \bf 11 & 0 \\
    \midrule
    Average & 58.07 & \underline{59.69} & 36.68  & \bf 61.19 & 41.21 \\
    \midrule
    TREC-Covid & \underline{7.37}  & 7.31 & 1.66 & \bf 11.68 & 4.48 \\
    NFCorpus & \underline{25.19} & \bf 29.88  & 24.57 & 25.03 & 23.22 \\
    Natural Questions & \bf 81.80 & \underline{80.45} & 44.02  & 75.11 & 43.47 \\
    HotpotQA & 60.05  & \underline{63.11} & 19.47 & \bf 64.67 & 41.63 \\
    FiQA & 51.67  & \underline{57.36} & 29.68 & \bf 60.85 & 32.21 \\
    ArguAna & 95.45 & \underline{96.66} & 83.86  & \bf 97.16 & 92.96 \\
    Tóuche-2020  & \underline{40.00} & 37.06  & 11.03 & \bf 47.19 & 10.05 \\
    CQAdupstack & 57.21 & \underline{59.76}  & 43.41 & \bf 64.72 & 45.54 \\
    Quora & \underline{97.97} & 97.59 & 93.69  & \bf 98.87 & 91.91 \\
    DBpedia & 40.39  & \underline{42.84} & 18.04 & \bf 45.63 & 23.79 \\
    Scidocs & 29.89 & \underline{33.02}  & 24.95 & \bf 34.80 & 25.37 \\
    Fever & \underline{90.84}  & 88.16 & 24.38 & \bf 92.20 & 25.70 \\
    Climate-Fever & 48.58 & \underline{52.29}  & 13.77 & \bf 52.71 & 32.65 \\
    Ccifact & \underline{86.60}  & \bf 90.17 & 81.06 & 86.00 & 83.97 \\
    \bottomrule
    \end{tabular}
    \caption{\label{BEIR_RESULTS_RECALL_COMPARE} Table comparing the performance of different systems on BEIR benchmark. All scores denote Recall@100. The best score is marked in bold and the second best is underlined. We evaluate BEIR in 2 different settings, zero-shot and unsupervised domain adaptation, being the latter the best method for maximum
performance.}
\end{table*}

\end{document}